\documentclass[letterpaper]{article} % DO NOT CHANGE THIS
\usepackage{aaai24}  % DO NOT CHANGE THIS
\usepackage{times}  % DO NOT CHANGE THIS
\usepackage{helvet}  % DO NOT CHANGE THIS
\usepackage{courier}  % DO NOT CHANGE THIS
\usepackage[hyphens]{url}  % DO NOT CHANGE THIS
\usepackage{graphicx} % DO NOT CHANGE THIS
\def\UrlFont{\rm}  % DO NOT CHANGE THIS
\usepackage{natbib}  % DO NOT CHANGE THIS AND DO NOT ADD ANY OPTIONS TO IT
\usepackage{caption} % DO NOT CHANGE THIS AND DO NOT ADD ANY OPTIONS TO IT
\usepackage{algorithm}
\usepackage{algorithmic}

\usepackage[
  detect-weight,
  exponent-product=\cdot,
  group-separator=.,
]{siunitx}
\DeclareSIUnit{\pp}{\textup{p.p.}}
\usepackage{import}
\usepackage{tikz}
\usepackage{booktabs}
\usepackage{pgfplots}
\usetikzlibrary{decorations.pathreplacing,calligraphy,backgrounds}
\usepackage{mathtools,amssymb}

\usepackage{newfloat}
\usepackage{listings}
\DeclareCaptionStyle{ruled}{labelfont=normalfont,labelsep=colon,strut=off} % DO NOT CHANGE THIS
\floatstyle{ruled}
\newfloat{listing}{tb}{lst}{}
\floatname{listing}{Listing}
\nocopyright
\newcommand{\netname}{MatSpectNet}

\title{MatSpectNet: Material Segmentation Network with Domain-Aware and Physically-Constrained Hyperspectral Reconstruction}

\author{
    Yuwen Heng\textsuperscript{\rm 1,\rm 2},
    Yihong Wu\textsuperscript{\rm 1},
    Jiawen Chen\textsuperscript{\rm 2},
    Srinandan Dasmahapatra\textsuperscript{\rm 1},
    Hansung Kim\textsuperscript{\rm 1}
}
\affiliations{
    \textsuperscript{\rm 1}School of Electronics and Computer Science \\ University of Southampton\\
     \textsuperscript{\rm 2}Baidu Inc.
\\Supplimentary material: https://github.com/heng-yuwen/MatSpectNet/blob/main/supplementary\%20material.pdf

}
\usepackage{bibentry}
\begin{document}

\maketitle

\begin{abstract}
  Achieving accurate material segmentation for 3-channel RGB images is challenging due to the considerable variation in a material's appearance. Hyperspectral images, which are sets of spectral measurements sampled at multiple wavelengths, theoretically offer distinct information for material identification, as variations in intensity of electromagnetic radiation reflected by a surface depend on the material composition of a scene. However, existing hyperspectral datasets are impoverished regarding the number of images and material categories for the dense material segmentation task, and collecting and annotating hyperspectral images with a spectral camera is prohibitively expensive. To address this, we propose a new model, the \netname{} to segment materials with recovered hyperspectral images from RGB images. The network leverages the principles of colour perception in modern cameras to constrain the reconstructed hyperspectral images and employs the domain adaptation method to generalise the hyperspectral reconstruction capability from a spectral recovery dataset to material segmentation datasets. The reconstructed hyperspectral images are further filtered using learned response curves and enhanced with human perception. The performance of \netname{} is evaluated on the LMD dataset as well as the OpenSurfaces dataset. Our experiments demonstrate that \netname{} attains a 1.60\% increase in average pixel accuracy and a 3.42\% improvement in mean class accuracy compared with the most recent publication. The project code is attached to the supplementary material and will be published on GitHub.
\end{abstract}

\section{Introduction}
\subimport{../figures}{overall_arch.tex}

Material segmentation aims to identify the material category of each pixel in the image. Although recent studies indicate that it is possible to achieve acceptable performance with annotated RGB datasets \cite{heng22camseg,Heng_2022_BMVC,schwartz2018visual,8675400,bell15minc}, the experiments in \cite{liang2022multimodal,mao2022surface} show that additional measurements of light such as near infra-red and laser beam reflection can distinguish materials more robustly. The theory is that the spectral profile of reflected electromagnetic waves is unique to various materials \cite{saragadam2020programmable,lichtman2005fluorescence,1990iii}. Since spectral cameras \cite{behmann2018specim} can capture the spectral profile of surface materials, it is feasible to use the hyperspectral images they produce for material segmentation.

While hyperspectral imaging has been widely used in geoscience and remote sensing \cite{zhong2016blind,kalman1997classification,li2022exploring,xue2021attention,mehta2021domain,liu2019review} over twenty years, the cost of collecting hyperspectral images hinders its widespread adoption in material segmentation for daily scenes \cite{stuart2022high}. A spectral camera can take a long acquisition time to scan a megapixel hyperspectral image with sufficient signal-noise ratio since the same amount of light has to be sampled at hundreds of wavelength bands \cite{behmann2018specim,zhang2019innovative}. This problem necessitates concessions in image spatial and spectral resolution. In addition, the ambient light should be able to cover the entire operating spectrum range, so the spectral camera should be used under daylight or halogen-based illumination. Before taking the hyperspectral images, the camera has to be calibrated with the measurement of black and white reference samples to analyse the material reliably \cite{behmann2018specim,shaikh2021calibration}. The stringent lighting requirements further restrict the application of hyperspectral images in indoor and motion scenes. 

In order to make spectral information more accessible for computer vision applications, researchers have been working on recovering spectral information from more easily obtainable data sources, such as RGB images \cite{arad2022ntire}. Over the past three years, several methods \cite{li2020deep,hu2022hdnet,cai2022mst++} have successfully improved the accuracy of reconstructed hyperspectral images. However, it remains unclear how these methods generalise to images captured by different camera models, as this aspect has not been explicitly investigated. In consideration of this problem, this paper proposes a novel Material Hyperspectral Network (\netname) to enhance the quality of recovered hyperspectral images on material datasets lacking RGB-hyperspectral image pairs. Figure \ref{overall} shows that the proposed \netname{} consists of two main sections. The network first learns to recover the hyperspectral images with the physical model of the camera, which serves as a constraint to ensure that the hyperspectral images preserve their physical property that the hyperspectral images can be transformed to RGB images with the physical camera model. Then the recovered hyperspectral images are processed with a multi-layer perceptron (MLP) to learn the material features from the spectral information at each pixel. 

To understand the proposed approach, it would be useful to delve into image theory first. An image is the quantitative measurement of the radiation from an illumination source or reflected by scene elements. Similar to the human perception system, the RGB camera measures the radiance of the visible spectrum with red, green and blue spectral response functions that accumulate the electromagnetic radiation from 380 to 720 nanometers and produce the raw-RGB values for each pixel \cite{magnusson2020creating}. The raw-RGB image is further processed with in-camera transformations including brightness adjustment and gamma correction to produce the final sRGB image, which is the format commonly used in vision datasets. 

Our proposed \netname{} model incorporates the physical relationship between hyperspectral and RGB images based on the image theory. Specifically, we exploit the fact that recovered hyperspectral images can be transformed into the original RGB counterparts through known spectral response functions and in-camera processing. This physical constraint is a key feature of our model and enables us to make reliable material predictions based on the recovered hyperspectral images.

%However, the accumulation effect can cause two different spectral distributions to be perceived as the same raw-RGB value, highlighting the disadvantage of relying on RGB images alone for material discrimination. On the other hand, the spectral camera directly measures the intensity of light usually sampled from 400 to 1000 nanometers \cite{arad2022ntire}, which covers both the visible and near-infrared spectrum to describe the optical properties of materials. Compared with RGB images, hyperspectral images contain more reliable information to distinguish materials. 

%The intensity is further transformed to reflectance relative to the black and white references. With the 

%With known RGB spectral response functions, it is possible to transform a hyperspectral image to its corresponding raw-RGB image \cite{magnusson2020creating}, which is also the main physical rule that we use in the \netname.

However, for open-access material segmentation datasets such as LMD \cite{8675400} and OpenSurfaces \cite{bell13opensurfaces}, the spectral response functions and in-camera image-processing pipeline are unknown. To bring hyperspectral images to the material segmentation task, the \netname{} models the physical camera with the sRGB transformation $R(h)$ that contains trainable components to adjust the unknown parameters. As illustrated in Figure \ref{overall}, given an sRGB image $x$, the \netname{} optimises that $R(S(x)) = x$ where $S(x)$ is the spectral recovery network that recovers hyperspectral images from sRGB ones. In practice, $S(x)$ is pre-trained on the ARAD\_1K dataset \cite{arad2022ntire} and fine-tuned together with material datasets. To align the data distribution, the idea of domain adaptation is used during training \cite{wu2021depth}. 

The recovered hyperspectral images are further processed with learned spectral response filters followed by a MLP to extract per-pixel material features. The spectral response filter is similar to the RGB spectral response functions in the mechanism, which aggregates the spectral information based on the sensitivity to the spectrum at each wavelength. The per-pixel material features are then tagged with the surface properties such as specularity and roughness queried from the most similar spectral measurement in the spectraldb dataset \cite{jakubiec2022data}, which serves as a piece of additional evidence to identify the materials.

The key contributions of this work are summarised as follows:
\begin{enumerate}
    \item \textit{Physically-Constrained Spectral Recovery.} Based on the theory that sRGB values can be obtained from hyperspectral images with known spectral response functions and in-camera processing, we propose to regulate the spectral recovery network with a trainable sRGB transformation.
    
    \item \textit{Domain-Aware Network Training.} To leverage the spectral recovery dataset for material segmentation, the domain adaptation is used to alleviate data distribution discrepancy between spectral recovery and material datasets. Moreover, domain-specific spectral response functions and image-processing pipelines are constructed. 
    
    \item \textit{Interpretable Hyperspectral Processing.} The learned spectral filters aggregate the spectra across the bandwidth and infer the electromagnetic frequency that contributes to material segmentation most.
    
    \item \textit{Multi-Modal Fused Material Segmentation.} The filtered per-pixel spectra and queried surface properties are fused together to make the material prediction from both spectral measurements as well as other empirical observations \cite{jain2013spectrophotometric,jakubiec2016building,jones2017experimental,lucas2014measuring}.

\end{enumerate}

The proposed network, \netname, outperforms existing models in the material segmentation task. With a per-pixel accuracy (Pixel Acc) of 88.24\% and a mean-class accuracy (Mean Acc) of 83.82\%, our network shows an improvement of 1.60\%/3.42\% over the most recent publication \cite{Heng_2022_BMVC}. Notably, \netname{} is particularly adept at recognising material categories that have limited samples, even under varying light conditions. These results are supported by per-category performance metrics and visualised segmentation results.

\section{Background}

\subsection{Material Segmentation}
The main challenge in material segmentation is the lack of reliable visual clues in RGB images. Recent studies have attempted to address this issue by forcing the network to learn material features from cropped image patches, assuming that the absence of object contours can prevent the network from converging to object characteristics such as shape \cite{heng22camseg,Heng_2022_BMVC,8675400,schwartz2016material}. However, these implicit methods are designed to learn local features that may improve material segmentation accuracy rather than extracting features that can accurately describe materials. Our proposed \netname{} takes a different approach by explicitly exploring material features from hyperspectral images, which quantitatively describe the reflection of light at tens of wavelengths for the surface materials. This approach can identify materials accurately, leading to improved segmentation performance. The justification for using hyperspectral images for material segmentation is in Section C.5 in the supplementary material.

\subsection{Material Property Measurements}
% Unlike the properties of an object, which are usually associated with common appearance semantics such as shape and colour, material properties define how the material interacts with light on its surface and shall be measured with certain measurement devices. For instance, the spectrophotometer \cite{albert2012low} hinges on the amount of light absorbed by the material to quantitatively measure the absorptance and reflectance distribution against visible and infrared radiation wavelength \cite{lv2022electromagnetic,van2014regular}. The streak camera \cite{bagayev2020experimental,horn2009ultra} measures the time-dependent temporal point spread functions (TPSF) \cite{kirkby1996measurement} which describes how light is reflected, refracted, scattered, or absorbed by the material. In addition to absorptance and reflectance, the femtosecond laser \cite{lureau2020high} which emits pulses of light can also measure the thermal conductivity and mechanical properties of the material by performing time-resolved measurements of temperature and laser-induced deformations \cite{guo2019ultrafast}. Although these cumbersome and costly machines can generate accurate material property measurements in the laboratory, the strict working environment makes them not suitable for outdoor and motion scenarios. 

Unlike the properties of objects, which are often associated with appearance semantics like shape and colour, the properties of materials are defined by how they interact with light at their surface and require specialised equipment for measurement. Portable measurement devices such as Time-of-Flight (ToF) cameras \cite{su2016material} and hyperspectral cameras \cite{behmann2018specim,s23031437} have the capability to assess the reflective or scattering properties of materials. ToF cameras operate as indirect sensors by determining the elapsed time for a light pulse to travel from the camera to the material and back \cite{su2016material}, while hyperspectral cameras directly capture the spectral signature, which quantifies how much light can be reflected by the material at sampled wavelengths. For material segmentation, hyperspectral cameras are preferred due to their ability to capture a complete scene and provide a comprehensive measurement of material properties through their spectral profiles. Additionally, the spectral profile can facilitate a correlation between measurements obtained from hyperspectral cameras and spectrophotometers (introduced in Section A in the supplementary material) provided that the measurements from the hyperspectral camera are lighting invariant.

In addition to sensor-based measurements, human perception also plays a crucial role in evaluating material properties. For instance, the photopic reflectance V($\lambda$) and melanopic reflectance M($\lambda$) are derived from the measured spectral profile based on the human visual system. The photopic reflectance captures the average human response to the brightness of light in the visible spectrum \cite{smith1996design}, while the melanopic reflectance provides information about the effect of reflected light on the activity of melanopsin photoreceptors in the human eye \cite{lucas2014measuring}. Moreover, the roughness or irregularity of material surfaces, which are difficult to measure with devices, can be estimated through appearance-driven assessments based on human observation \cite{jakubiec2022data,jones2017experimental}.

In this research, we employ the ARAD\_1K dataset \cite{arad2022ntire}, captured by a hyperspectral camera, as the training data for our recovery network, $S(x)$. To refine the precision of the recovered hyperspectral images and incorporate human observations, we utilise the spectral and observation measurements in the spectraldb \cite{jakubiec2022data}, acquired from a spectrophotometer, as a correction reference.

\subsection{Hyperspectral Image Recovery Methods}
Early attempts to recover hyperspectral images rely on sparse coding methods, such as manifold representation, which embeds high-dimensional spectral information into low-dimensional representations \cite{li2020deep,jia2017rgb}. Recent network-based methods investigate network modules that learn both spatial and spectral features \cite{cai2022mst++,hu2022hdnet}. While these methods have achieved accurate spectral recoveries, the application of hyperspectral images is still limited by the lack of annotated hyperspectral image datasets with semantic labels. Despite advancements in solving the spectral recovery challenge for geoscience applications such as aerial image dehazing, the challenge remains a topic of ongoing research for daily images \cite{mehta2021domain,cai2022coarse,arad2022ntire}. In this paper, we investigate how to apply hyperspectral recovery methods to existing material segmentation datasets. The proposed method can also be applied to other tasks without much modification.

\section{Methodology}
The proposed \netname{} aims to tackle the material segmentation task by utilising the discriminatory information from reconstructed hyperspectral images. However, the lack of open-access datasets with corresponding hyperspectral images presents an additional challenge for the material segmentation task. To address this, we introduce a physically-constrained spectral recovery architecture and employ a domain-aware training approach (illustrated in Figure \ref{overall}, (a)) to incorporate hyperspectral information into the material segmentation task. The recovered hyperspectral images are then processed using interpretable spectral filters and combined with other observations such as roughness to improve the segmentation quality (shown in Figure \ref{overall}, (b)).

\subsection{Physically-Constrained Spectral Recovery}
\label{sec.pcsp}
\subimport{../figures}{spectral_recovery_net_archv2.tex}

The spectral recovery network learns the transformation from an sRGB image $x \in \mathcal{X}$ to its corresponding hyperspectral image $\hat{h}\in \mathcal{H}$ via $S:\mathcal{X}\to \mathcal{H}$ \cite{cai2022mst++,li2022drcr,agarla2022fast,arad2022ntire}. However, evaluating the quality of reconstructed hyperspectral images is challenging due to the absence of measured hyperspectral images in material datasets. To address this issue, we introduce an RGB transformation network $R:\mathcal{H} \to \mathcal{X}$ that maps the reconstructed hyperspectral image $\hat{h}$ back to the corresponding sRGB image, as shown in Figure \ref{fig:reverser}. We select the best $S^*_\theta(x) = \arg\min_{S_\theta(x)} L_{trans}$ by monitoring the loss term for the material datasets:
\begin{equation}
    L_{trans} = L_{MSE}(x, R(S_\theta(x)))
\end{equation}
To recover accurate hyperspectral images for material datasets by minimizing the transformation loss, the design of $R(h)$ must reflect the physical relationship of the RGB and hyperspectral image pair. Hence, a simple network, as used in previous works such as \cite{mehta2021domain}, is not sufficient. To address this, $R(h)$ explicitly incorporates the physical RGB camera model including response functions, brightness, and system noise, as shown in Equation \ref{eq:rx}:
\begin{equation}
R(h) = f_{jpeg}(\mu f_{noise}(W_{rgb}h))
\label{eq:rx}
\end{equation}
\noindent where the hyperspectral image $h \in \mathbb{R}^{n\_bands\times H \times W}$ and the RGB response functions are formatted as a matrix $W_{rgb} \in \mathbb{R}^{3\times n\_bands}$. The camera noise is included by the function $f_{noise}$, and $\mu$ is the brightness factor. Here, $n\_bands$ is the number of spectra bands sampled by the response functions. The function $f_{jpeg}$ models the in-camera processing and compression noise introduced by the JPEG compression algorithm. All components are trainable in Figure \ref{fig:reverser} and will be explained in the following paragraphs. This simplified camera model is validated in Section C in the supplementary material.

\subsubsection{RGB Response Functions.} All digital light sensors exhibit varying sensitivity to different wavelength ranges of light due to their spectral response functions \cite{tropp2017mathematical}. Specifically, trichromatic cameras or three-colour image sensors, inspired by human colour perception, have unique spectral response functions in their red (R), green (G), and blue (B) channels based on the tristimulus theory \cite{smithson2005sensory}. That is to say, if we know the measurement of the spectral reflection  $h$ and the response matrix $W_{rgb}$, sampled at the same bands, the noiseless RGB image $rgb_{clean}$ can be obtained as $W_{rgb}h$, on the condition that $h$ is properly calibrated \cite{ji2021compressive,behmann2018specim}. However, the RGB values are not consistent across various camera models, even under the same capturing conditions. This is because each model of three-colour image sensors responds differently to light due to their unique RGB spectral response functions. Hence, the RGB values are device-specific and not interchangeable among different camera models.

In order to handle images taken by different camera models \cite{8675400,bell15minc,bell13opensurfaces}, we propose to learn the sensitivity displacement $\Delta band_i$ at each spectral band $i$ compared with standard response functions based on the input sRGB image, as depicted in Figure 10 in the supplementary material. The standard response curves are sourced from the ARAD\_1K dataset, which is based on physical measurements of a Basler Ace 2 camera \cite{arad2022ntire}. To learn the sensitivity displacement, we redesigned the spectral recovery network $S(x)$, which obeys the encoder-decoder architecture, by attaching one auxiliary path to the encoder. As shown in Figure \ref{fig:response_shift}, the encoder output is processed with the repetitive spectra processing module comprising 1$\times$1 convolutional kernels and average pooling. The 1$\times$1 convolutional kernel is responsible to learn from the channel information since the band difference is applied to each pixel individually. The average pooling downsamples the feature map to correct the prediction. Moreover, the response curves should learn from long, middle, and short wavelength regions for R, G, and B channels respectively. To keep maintain the functionality of the response curves, the band differences are aggregated as a loss term $L_{band}$:
\begin{equation}
\label{eq:lband}
    L_{band} = \sum_{r,g,b} \sum_\lambda band_{\lambda|r,g,b} \times ||\Delta band_{\lambda|r,g,b}||
\end{equation}
where $band_\lambda$ is sampled from the standard response curve at wavelength $\lambda$ for channel R or G or B. In this way, the highly sensitive region incurs severe penalties, causing its displacement $\Delta band_\lambda$ to be zero and preserving its functionality. In our experiment, the MST++\cite{cai2022mst++} is chosen to be our spectral recovery network $S(x)$ and the spectra processing module in Figure \ref{fig:response_shift} is repeated three times with channel numbers 124, 62, 31+3 where 31 of them is the number of spectra band sampled between 400nm and 700nm with step 10nm, and the other 3 scalars are the noise parameters and brightness factor explained in following sections. 
\subimport{../figures}{spectral_shift_net.tex}
\subsubsection{Camera System Noise and Brightness.} Since the hyperspectral image $h$ in the ARAD\_1K datasaet is calibrated with white and dark reference samples to measure the actual reflectance rather than signal intensity, the recovered hyperspectral image $\hat{h}$ is projected into noiseless RGB $rgb_{clean}$ by $W_{rgb}\hat{h}$. However, for realistic cameras, the camera system noise caused by unwanted variations in the signals produced by the image sensor and processing circuitry can reduce the image quality, particularly in low light conditions or high ISO settings \cite{baxter2012calibration,shin2019camera,park2020median}. Camera system noise can be categorized into several types, including thermal noise caused by random fluctuations in the electrical charge generated by the image sensor due to heat \cite{berthelon2018effects}, and shot noise caused by the random nature of the way light interacts with the image sensor \cite{roussel2018polarimetric}. To address the camera system noise, we simulate camera system noise using parameterised probability models. 

Since thermal noise is a form of Gaussian noise, it can be modelled with zero-mean normal distribution $N(0, \sigma)$ where the standard variance $\sigma$ is considered as the noise level \cite{denk2007modelling,chen2021thermal}. As for the shot noise, it arises due to the random nature of the arrival times of particles such as photons at the sensor, hence it can be modelled with Poisson distribution \cite{roussel2018polarimetric,arad2022ntire} $P(\nu)$ where $\nu$ is the noise level which is proportional to the intensity of the incoming light. In summary, the noisy RGB can be represented as:
\begin{equation}
    rgb_{noisy} = \mu P((N(0, \sigma) + rgb_{clean})\times\nu) /\nu
\end{equation}
where the brightness factor $\mu$ adjusts the intensity of the image with the average scene brightness assumption. The noise level $\nu$, $\sigma$ are tuned and justified in Section C.2 in the supplementary material. In practice, the uniform brightness factor is redundant when the ground truth images and recovered images are normalised into the range [0,1]. The comparison between using [0,1] normalisation and using brightness factor is in Section C.1 in the supplementary material.  

\subsubsection{Other Components and Compression Noise.}
In a typical in-camera processing pipeline, the final sRGB image is derived through various processing steps from the initial noisy raw image $rgb_{noisy}$. These steps typically involve operations such as white balance, colour manipulation \citep{tseng2022neural}, gamma correction, and JPEG compression \citep{prakash2022study}. While the sRGB gamma correction follows a known equation and JPEG compression can be approximated mathematically with differentiable operations \citep{wang2022neural,mishra2022deep,shin2017jpeg}, the specific image style configurations for colour manipulation vary across different camera models and are often intricate and proprietary. To avoid the explicit modelling of these camera-specific components applied to the noisy RGB images, this section proposes a network that encompasses the effects of in-camera processing and the noise induced by JPEG compression. This approach is achieved through the integration of two basic Swin transformer layers \citep{liu2021swin, liu2021swin2} to post-process the noisy image. The basic Swin layer consists of a window-based self-attention and MLP processing to learn from local regions. The window size is set to 8, as the JPEG compression algorithm typically applies the 2D discrete cosine transform on 8×8 blocks. This framework enables the generation of final sRGB images that faithfully capture the effects of these in-camera processes, providing a more robust representation of the recovered sRGB images. Though the network architecture may sound simple, the experiments in Section C.3 in the supplementary material show that a simple network is enough to cope with the in-camera processing and compression noise.
\subsection{Domain-Aware Network Training}
\label{sec.domain}
\begin{figure}
    \centering

\resizebox{\columnwidth}{!}{%
\tikzset{every picture/.style={line width=0.75pt}} %set default line width to 0.75pt        

\begin{tikzpicture}[x=0.75pt,y=0.75pt,yscale=-1,xscale=1]
%uncomment if require: \path (0,396); %set diagram left start at 0, and has height of 396

%Straight Lines [id:da29828265219945016] 
\draw [line width=1.5]    (67.71,55.5) -- (107,55.5) ;
\draw [shift={(110.96,55.5)}, rotate = 180] [fill={rgb, 255:red, 0; green, 0; blue, 0 }  ][line width=0.08]  [draw opacity=0] (11.61,-5.58) -- (0,0) -- (11.61,5.58) -- cycle    ;
%Straight Lines [id:da10520982939383772] 
\draw [line width=1.5]    (189.71,55.5) -- (228.96,55.5) ;
\draw [shift={(232.96,55.5)}, rotate = 180] [fill={rgb, 255:red, 0; green, 0; blue, 0 }  ][line width=0.08]  [draw opacity=0] (11.61,-5.58) -- (0,0) -- (11.61,5.58) -- cycle    ;
%Rounded Rect [id:dp5129992308143838] 
\draw  [fill={rgb, 255:red, 248; green, 231; blue, 28 }  ,fill opacity=1 ] (116.71,100.88) .. controls (116.71,96.46) and (120.29,92.88) .. (124.71,92.88) -- (178.71,92.88) .. controls (183.13,92.88) and (186.71,96.46) .. (186.71,100.88) -- (186.71,124.88) .. controls (186.71,129.3) and (183.13,132.88) .. (178.71,132.88) -- (124.71,132.88) .. controls (120.29,132.88) and (116.71,129.3) .. (116.71,124.88) -- cycle ;

%Rounded Rect [id:dp38386542910444854] 
\draw  [fill={rgb, 255:red, 74; green, 144; blue, 226 }  ,fill opacity=1 ] (115.71,42.88) .. controls (115.71,38.46) and (119.29,34.88) .. (123.71,34.88) -- (177.71,34.88) .. controls (182.13,34.88) and (185.71,38.46) .. (185.71,42.88) -- (185.71,66.88) .. controls (185.71,71.3) and (182.13,74.88) .. (177.71,74.88) -- (123.71,74.88) .. controls (119.29,74.88) and (115.71,71.3) .. (115.71,66.88) -- cycle ;

%Straight Lines [id:da11831517996596674] 
\draw [line width=1.5]    (195.71,112.5) -- (234.96,112.5) ;
\draw [shift={(191.71,112.5)}, rotate = 0] [fill={rgb, 255:red, 0; green, 0; blue, 0 }  ][line width=0.08]  [draw opacity=0] (11.61,-5.58) -- (0,0) -- (11.61,5.58) -- cycle    ;
%Straight Lines [id:da6058198254043183] 
\draw [line width=1.5]    (73.71,112.5) -- (112.96,112.5) ;
\draw [shift={(69.71,112.5)}, rotate = 0] [fill={rgb, 255:red, 0; green, 0; blue, 0 }  ][line width=0.08]  [draw opacity=0] (11.61,-5.58) -- (0,0) -- (11.61,5.58) -- cycle    ;
%Shape: Rectangle [id:dp4205890718971843] 
\draw  [dash pattern={on 4.5pt off 4.5pt}] (30,30) -- (271.13,30) -- (271.13,138.05) -- (30,138.05) -- cycle ;
%Straight Lines [id:da17107607370134525] 
\draw [line width=1.5]    (322.71,55.5) -- (361.96,55.5) ;
\draw [shift={(365.96,55.5)}, rotate = 180] [fill={rgb, 255:red, 0; green, 0; blue, 0 }  ][line width=0.08]  [draw opacity=0] (11.61,-5.58) -- (0,0) -- (11.61,5.58) -- cycle    ;
%Straight Lines [id:da08689941508124832] 
\draw [line width=1.5]    (444.71,55.5) -- (483.96,55.5) ;
\draw [shift={(487.96,55.5)}, rotate = 180] [fill={rgb, 255:red, 0; green, 0; blue, 0 }  ][line width=0.08]  [draw opacity=0] (11.61,-5.58) -- (0,0) -- (11.61,5.58) -- cycle    ;
%Rounded Rect [id:dp44276531653832873] 
\draw  [fill={rgb, 255:red, 248; green, 231; blue, 28 }  ,fill opacity=1 ] (371.71,100.88) .. controls (371.71,96.46) and (375.29,92.88) .. (379.71,92.88) -- (433.71,92.88) .. controls (438.13,92.88) and (441.71,96.46) .. (441.71,100.88) -- (441.71,124.88) .. controls (441.71,129.3) and (438.13,132.88) .. (433.71,132.88) -- (379.71,132.88) .. controls (375.29,132.88) and (371.71,129.3) .. (371.71,124.88) -- cycle ;

%Rounded Rect [id:dp44748162113844536] 
\draw  [fill={rgb, 255:red, 74; green, 144; blue, 226 }  ,fill opacity=1 ] (370.71,42.88) .. controls (370.71,38.46) and (374.29,34.88) .. (378.71,34.88) -- (432.71,34.88) .. controls (437.13,34.88) and (440.71,38.46) .. (440.71,42.88) -- (440.71,66.88) .. controls (440.71,71.3) and (437.13,74.88) .. (432.71,74.88) -- (378.71,74.88) .. controls (374.29,74.88) and (370.71,71.3) .. (370.71,66.88) -- cycle ;

%Straight Lines [id:da8174954703096746] 
\draw [line width=1.5]    (450.71,112.5) -- (489.96,112.5) ;
\draw [shift={(446.71,112.5)}, rotate = 0] [fill={rgb, 255:red, 0; green, 0; blue, 0 }  ][line width=0.08]  [draw opacity=0] (11.61,-5.58) -- (0,0) -- (11.61,5.58) -- cycle    ;
%Straight Lines [id:da5318976568121268] 
\draw [line width=1.5]    (328.71,112.5) -- (367.96,112.5) ;
\draw [shift={(324.71,112.5)}, rotate = 0] [fill={rgb, 255:red, 0; green, 0; blue, 0 }  ][line width=0.08]  [draw opacity=0] (11.61,-5.58) -- (0,0) -- (11.61,5.58) -- cycle    ;
%Shape: Rectangle [id:dp1606854723351141] 
\draw  [dash pattern={on 4.5pt off 4.5pt}] (285,30) -- (526.13,30) -- (526.13,138.05) -- (285,138.05) -- cycle ;

% Text Node
\draw (134,104.38) node [anchor=north west][inner sep=0.75pt]   [align=left] {$R_s(h)$};
% Text Node
\draw (135.71,47.5) node [anchor=north west][inner sep=0.75pt]   [align=left] {$S(x)$};
% Text Node
\draw (40.46,50) node [anchor=north west][inner sep=0.75pt] [font=\Large]   {$x_s$};
% Text Node
\draw (240,42) node [anchor=north west][inner sep=0.75pt]  [font=\Large]  {$\hat{h}_s$};
% Text Node
\draw (240,100) node [anchor=north west][inner sep=0.75pt]  [font=\Large]  {$\hat{h}_s$};
% Text Node
\draw (41,102) node [anchor=north west][inner sep=0.75pt]  [font=\Large]   {$\hat{x}_s$};
% Text Node
\draw (294.46,50) node [anchor=north west][inner sep=0.75pt]  [font=\Large]   {$x_m$};
% Text Node
\draw (494.46,42) node [anchor=north west][inner sep=0.75pt]  [font=\Large] {$\hat{h}_m$};
% Text Node
\draw (495.46,100) node [anchor=north west][inner sep=0.75pt]  [font=\Large]  {$\hat{h}_m$};
% Text Node
\draw (295.46,101) node [anchor=north west][inner sep=0.75pt]  [font=\Large]  {$\hat{x}_m$};
% Text Node
\draw (390.71,47.5) node [anchor=north west][inner sep=0.75pt]   [align=left] {$S(x)$};
% Text Node
\draw (387,104.38) node [anchor=north west][inner sep=0.75pt]   [align=left] {$R_m(h)$};
% Text Node
% \draw (129,146) node [anchor=north west][inner sep=0.75pt]   [align=left] {W\_rgb};
% % Text Node
% \draw (382,146) node [anchor=north west][inner sep=0.75pt]   [align=left] {W\_rgb\textasciicircum *};
% Text Node
\draw (31,10) node [anchor=north west][inner sep=0.75pt]   [align=left] [font=\large] {\textbf{spectral dataset}};
% Text Node
\draw (286,10) node [anchor=north west][inner sep=0.75pt]   [align=left] [font=\large] {\textbf{material dataset}};

%%%%%%%%%%%%%%%%%%%%%%%%
%Straight Lines [id:da5268163498713656] 
\draw [line width=1.5]    (605.71,56.38) -- (644.96,56.75) ;
\draw [shift={(648.96,56.79)}, rotate = 180.54] [fill={rgb, 255:red, 0; green, 0; blue, 0 }  ][line width=0.08]  [draw opacity=0] (11.61,-5.58) -- (0,0) -- (11.61,5.58) -- cycle    ;
%Straight Lines [id:da79335232538129] 
\draw [line width=1.5]    (727.71,55.38) -- (750.96,55.75) ;
\draw [shift={(751.96,55.79)}, rotate = 180.54] [fill={rgb, 255:red, 0; green, 0; blue, 0 }  ][line width=0.08]  [draw opacity=0] (11.61,-5.58) -- (0,0) -- (11.61,5.58) -- cycle    ;
%Rounded Rect [id:dp9434390080369051] 
\draw  [fill={rgb, 255:red, 74; green, 144; blue, 226 }  ,fill opacity=1 ] (654.71,100.88) .. controls (654.71,96.46) and (658.29,92.88) .. (662.71,92.88) -- (716.71,92.88) .. controls (721.13,92.88) and (724.71,96.46) .. (724.71,100.88) -- (724.71,124.88) .. controls (724.71,129.3) and (721.13,132.88) .. (716.71,132.88) -- (662.71,132.88) .. controls (658.29,132.88) and (654.71,129.3) .. (654.71,124.88) -- cycle ;

%Rounded Rect [id:dp8792109878731198] 
\draw  [fill={rgb, 255:red, 248; green, 231; blue, 28 }  ,fill opacity=1 ] (653.71,43.88) .. controls (653.71,39.46) and (657.29,35.88) .. (661.71,35.88) -- (715.71,35.88) .. controls (720.13,35.88) and (723.71,39.46) .. (723.71,43.88) -- (723.71,67.88) .. controls (723.71,72.3) and (720.13,75.88) .. (715.71,75.88) -- (661.71,75.88) .. controls (657.29,75.88) and (653.71,72.3) .. (653.71,67.88) -- cycle ;

%Straight Lines [id:da5282897073488892] 
\draw [line width=1.5]    (733.71,112.42) -- (752.96,112.79) ;
\draw [shift={(729.71,112.38)}, rotate = 0.54] [fill={rgb, 255:red, 0; green, 0; blue, 0 }  ][line width=0.08]  [draw opacity=0] (11.61,-5.58) -- (0,0) -- (11.61,5.58) -- cycle    ;
%Straight Lines [id:da8525152769243973] 
\draw [line width=1.5]    (633,114.42) -- (650.96,114.79) ;
\draw [shift={(632,114.38)}, rotate = 0.54] [fill={rgb, 255:red, 0; green, 0; blue, 0 }  ][line width=0.08]  [draw opacity=0] (11.61,-5.58) -- (0,0) -- (11.61,5.58) -- cycle    ;
%Shape: Rectangle [id:dp6508620956979398] 
\draw  [dash pattern={on 4.5pt off 4.5pt}] (540,30) -- (830,30) -- (830,138.05) -- (540,138.05) -- cycle ;

% Text Node
\draw (570,46) node [anchor=north west][inner sep=0.75pt] [font=\Large]  [align=left] {$h_s$};
% Text Node
\draw (752,46) node [anchor=north west][inner sep=0.75pt] [font=\Large]  [align=left] {$R_s(h_s)$};
% Text Node
\draw (752,103.88) node [anchor=north west][inner sep=0.75pt]  [font=\Large] [align=left] {$R_s(h_s)$};
% Text Node
\draw (550.46,103.88) node [anchor=north west][inner sep=0.75pt] [font=\Large]  [align=left] {$S(R_s(h_s))$};

% Text Node
\draw (541,10) node [anchor=north west][inner sep=0.75pt]   [align=left] {\textbf{spectral dataset} (measured $h$ as input)};
% Text Node
\draw (670,47) node [anchor=north west][inner sep=0.75pt]   [align=left] {$R_s(h)$};
% Text Node
\draw (673,104.5) node [anchor=north west][inner sep=0.75pt]   [align=left] {$S(x)$};

\end{tikzpicture}

}
    
    \caption{The data flow for spectral as well as material datasets. Specifically, these two datasets share the same $S(x)$ but have their own $R(h)$ to model the cameras.}
    \label{fig:train_rgb_fig}
\end{figure}
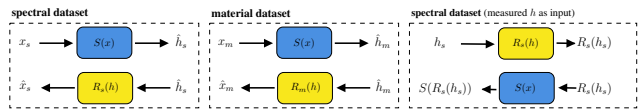
The goal of this paper is to employ hyperspectral images for material segmentation. As there are no hyperspectral images in material datasets, we propose to incorporate the spectral recovery dataset ARAD\_1K \cite{arad2022ntire}, which includes pairs of RGB and hyperspectral images, to jointly train the spectral recovery network $S(x)$ and the RGB transformation network $R(h)$ with the material datasets. We denote the samples in the spectral dataset as ($x_s$,$h_s$), and the samples in the material dataset as ($x_m$). The total loss is defined as $L_{total} = 10\times L_{band} + 5\times (L_{rgb} + L_{spectral}) + 0.5\times L_{domain}$. Equation \ref{eq:lband} has introduced the band loss, while the remaining three terms will be introduced in this section. The domain loss $L_{domain}$ is introduced in Section B.1 in the supplementary material.

\subsubsection{RGB Recovery Loss.}
Ideally, an RGB image passing through $S(x)$ and $R(h)$ should be able to recover itself, as shown in Equation \ref{eq:rgbloss}. 
\begin{equation}
\begin{split}
\MoveEqLeft
    L_{rgb} =  L_{trans} + L_{MSE}(x_s, \hat{x}_s)  \\
    &+ L_{MSE}(x_s, R_s(h_s))
    \label{eq:rgbloss}
\end{split}
\end{equation}
where $\hat{x}_s$, $\hat{x}_m$ are recovered RGB images, $R_s(h_s)$ takes the measured hyperspectral image as input, and $L_{MSE}$ is the Mean Squared Error. As shown in Figure \ref{fig:train_rgb_fig}, $x_s$ and $x_m$ share the same spectral recovery network $S(x)$. As for the RGB transformation networks $R_s(h)$ and $R_m(h)$, these two datasets have their own trainable parameters since the RGB images are taken by different cameras.

\subsubsection{Spectral Recovery Loss.} To calculate the spectral recovery loss term $L_{spectral}$, we compare the recovered hyperspectral image $\hat{h}_s$ with its ground truth $h_s$, using the mean relative absolute error ($L_{MRAE}$) \cite{arad2022ntire}:
\begin{equation}
    L_{spectral} = L_{MRAE}(h_s, \hat{h}_s) + L_{MRAE}(h_s, S(R_s(h_s)))
\end{equation}
The two terms correspond to the left and right parts of Figure \ref{fig:train_rgb_fig}. The second term $S(R_s(h_s))$ reverses the order of operations in Figure \ref{fig:train_rgb_fig} to ensure that the networks $S(x)$ and $R(h)$ are order-irrelevant. Moreover, this term also ensures that the recovered RGB image can be successfully transformed back to the original hyperspectral image.

\subsection{Interpretable Hyperspectral Processing}
\label{sec.filter}
The recovery of hyperspectral images opens up the possibility of understanding the contribution of each wavelength to the material segmentation task. This information is useful for modifying camera response curves to generate task-specific images \cite{saragadam2020programmable}. To facilitate this, we propose a wavelength-wise self-attention module, named \textquotesingle spectral attention\textquotesingle, which processes the hyperspectral images using predicted filters that have a similar physical meaning as RGB response curves, as shown in Figure \ref{fig:spectral_attn}. By aggregating the hyperspectral images based on their dependencies at each wavelength, this module can identify the spectral information that is most relevant to the task at hand. More detail is in Section D.2 in the supplementary material.

\subimport{../figures}{spectral_attention.tex}
\subsection{Multi-Modal Fusion}
\label{sec.fusion}

In addition to the spectral information learned from the ARAD\_1K dataset, we also want to incorporate human material observations from the spectraldb dataset into the segmentation process. To query the observations, we use the per-pixel spectral measurements $s$ as the bridge to link these two datasets. It is worth noting that the spectral camera and spectrophotometer have different measurement precisions, despite measuring the same physical property. Therefore, in this work, we compare the difference with the spectra shape matrix $S$ of the spectra measurements. The matrix element $S_{\lambda_a,\lambda_b}$ is defined in Equation \ref{shape_diff}:
\begin{equation}
    S_{\lambda_a,\lambda_b} = |s_{\lambda_a} - s_{\lambda_b}|
\label{shape_diff}
\end{equation}
where $\lambda_a$ and $\lambda_b$ are the wavelength bands within the range [400nm, 700nm]. We construct the spectra shape matrix for both the recovered hyperspectral images as well as the spectraldb measurements \cite{jakubiec2022data}, which contains multiple spectra measurements indexed by $k$.

As shown in Figure \ref{fig:align_spectral}, for each pixel $i,j$ in the hyperspectral image, we find the matched measurement  $S_k^*=\arg\min_{S_k}||S[i,j]- S_k||_2$ with the $L_2$ distance. Then the corresponding observations including reflectance, specularity and roughness are appended to pixel $i,j$ of the recovered hyperspectral image. The queried observations together with the filtered features are passed into MLP to extract material features, as shown in Figure \ref{overall}. The way we generate the final material prediction is described in Section B.5 in the supplementary material.

\subimport{../figures}{align_spectral.tex}

% Remember to use \url{https://www.specim.com/downloads/iq/manual/software/iq/topics/data-cube.html} to deduce the rescale operation.

\begin{table*}
  \centering
\resizebox{1.6\columnwidth}{!}{%
\begin{tabular}{ c|cc|ccc|ccc} 
  \toprule
  Datasets & \multicolumn{2}{c|}{LMD}  & \multicolumn{3}{c|}{OpenSurfaces } & \multicolumn{3}{c}{--}  \\
  Architecture & Pixel Acc (\%) & Mean Acc (\%) & Pixel Acc (\%) & Mean Acc (\%) & mIoU (\%)	& \#params (M) & \#flops (G) & FPS \\
   \midrule 
ResNeSt-101  \cite{zhang2022resnest}      & 82.45 $\pm$ 0.20    & 75.31 $\pm$ 0.29          & 85.10 & 67.13 & 55.32 & 48.84 & 63.39 &25.57\\
EfficientNet-b5 \cite{tan2019efficientnet}   &   83.17 $\pm$ 0.06        &  76.91 $\pm$ 0.06         &  84.63 & 65.47   & 53.25  &  30.17 & 20.5 &27.00\\

Swin-t   \cite{liu2021swin}          &  84.70 $\pm$ 0.26       & 79.06 $\pm$ 0.46         & 86.19 & 69.41 &  57.71     &29.52 & 34.25&33.94\\
CAM-SegNet-DBA \cite{Heng_2022_BMVC} & 86.12 $\pm$ 0.15 & 79.85 $\pm$ 0.28 & 86.64 & 69.92 & 58.18 & 68.58 & 60.83 & 17.79\\
%Swin-S            &   87.18         &    82.09       &      &        &         &  50.83    & 58.02 & 25.46 \\
DBAT   \cite{Heng_2022_BMVC}         & 86.85 $\pm$ 0.08 & 81.05 $\pm$ 0.28 & 86.28 & 70.68 &  58.08   & 56.03 & 41.23&27.44\\

\textbf{MatSpectNet}    & \textbf{88.24 $\pm$ 0.10} & \textbf{83.82 $\pm$ 0.23} & \textbf{87.13} & \textbf{71.29} & \textbf{58.92} & 57.7 & 42.16 & 26.83 \\

 \bottomrule
 \end{tabular}
 }
%   \vspace{0.2pt}
  \caption{The performance \cite{heng22camseg} reported on the LMD and the OpenSurfaces. The FPS value of \netname{} is calculated by processing 1000 images with one NVIDIA 3090. The uncertainty evaluation is reported across five trainings.}
  \label{testacc}
\end{table*}

\section{Experiments}
In this paper, we assess the performance of our network on two material datasets, namely the LMD \cite{schwartz2018visual,8675400} and the OpenSurfaces \cite{bell13opensurfaces}, using the evaluation metrics reported in \cite{Heng_2022_BMVC}. The \netname{} is trained for 400 epochs with 8 NVIDIA GeForce RTX 3090 GPUs. The training configurations are in Section B, the spectral recovery experiments are in Section C, and the additional analysis is in Section D in the supplementary material.

\subsection{Quantitative Evaluation}
Table \ref{testacc} presents numerical evaluations of seven networks. Our \netname{} outperforms all other compared networks on both datasets. Specifically, on the LMD, our \netname{} achieves the highest Pixel Acc of 88.24\% and Mean Acc of 83.82\%. Compared to the second-best performing network, DBAT \cite{Heng_2022_BMVC}, the \netname{} shows improvements of 1.60\% and 3.42\% for Pixel Acc and Mean Acc, respectively. On the OpenSurfaces dataset, the \netname{} also achieves the best performance with accuracy scores of 87.13\% and 71.29\% for Pixel Acc and Mean Acc, respectively. The corresponding improvements compared to the second-best scores are 0.99\% and 0.86\%. In particular, for the LMD dataset, the higher improvement in Mean Acc and lower improvement in Pixel Acc indicate that our \netname{} improves the performance for hard-to-recognize material categories compared to other networks, suggesting that hyperspectral information can provide reliable material features. The detailed per-category performance is reported in Section D.1 in the supplementary material. 

\subsection{Qualitative Evaluation}
\subimport{../figures}{qualitive_vis.tex}
The segmented images are displayed in Figure \ref{B-4.1}. Despite the sparsely labelled ground truth \cite{heng22camseg,Heng_2022_BMVC,8675400}, our \netname{} is the only model that can accurately classify nearly all the pixels belonging to the stone wall, even those under bright lighting conditions. This suggests that the calibrated spectral information can boost network performance under varying illumination conditions, as compared to other purely image-based networks.

However, we also noticed that the material labels defined in the LMD dataset lack the ability to fully describe indoor scenes in a dense manner. Furthermore, the similarity in appearance of different materials makes the segmentation task challenging to solve with RGB images. As spectral information is independent of visual appearance, we believe that hyperspectral reconstruction could be a potential solution to overcome the limitation of sparse labeling for pixels that cannot be adequately described by the given labels.

\section{Ablation Study}
\begin{table}
  \centering
\resizebox{0.9\columnwidth}{!}{%
\begin{tabular}{ c|ccc} 
  \toprule
Methods   & Regularised Spectral Recovery & Domain Alignment  & Multi-Modal Fusion  \\
   \midrule 
Network 1 & & & \\ 
Network 2 & \checkmark & & \\ 
Network 3 & \checkmark & \checkmark & \\ 
\netname & \checkmark & \checkmark & \checkmark \\ 
 \bottomrule
 \end{tabular}
 }
%   \vspace{0.2pt}
  \caption{The network variations and the corresponding enabled components.}
  \label{tune_components}
\end{table}
\begin{table}
  \centering
\resizebox{0.9\columnwidth}{!}{%
\begin{tabular}{ c|cccc} 
  \toprule
  n\_filters  & 4 & 8 & 12  & 16  \\
   \midrule 

% Pixel Acc & 87.55 & 86.95 (+0.63) & 86.72 (+0.25) & 87.34 & 87.10 & 87.17  & 87.76 & 86.63 \\ 
% Mean Acc & 83.90 & 82.82 (+1.21) & 83.02 (-0.01) & 83.49 & 82.92 & 83.51 & 84.10 & 82.48 \\ 
% Mean IoU & 72.63 & 71.15 (+1.81) & 71.32 (+0.33) & 72.11 & 71.47 & 72.38 & 73.57 & 71.25 \\

Network 1 & 86.73/80.91 & 86.55/80.14 & 86.94/81.13 & 87.08/81.20 \\
Network 2 & 86.93/81.41 & 87.17/81.74 & 87.69/81.92 & 87.10/81.63 \\
Network 3 & 87.16/82.88 & 87.22/82.93 & 87.46/81.71 & 87.54/82.36 \\
\netname  & 87.95/83.17 & 87.66/83.14 & \textbf{88.24/83.82} & 88.07/83.52 \\

 \bottomrule
 \end{tabular}
 }
%   \vspace{0.2pt}
  \caption{The network performance for each variation against four filter number choices.}
  \label{tune_filter}
\end{table}
In this section, we assess the impact of each component of the proposed \netname{} by introducing three additional models of increasing complexity, as shown in Table \ref{tune_components}. For each model, we adjust the number of filters in the interpretable hyperspectral processing module and report the corresponding (Pixel Acc/Mean Acc) results on the LMD dataset in Table \ref{tune_filter}.

\subsubsection{Regularised Spectral Recovery.} Network 1 utilises a pre-trained spectral recovery network to generate the hyperspectral images. However, since the network is not specifically tuned for the material datasets, the network trained on the ARAD\_1K dataset is not generalisable, and the network performance is hardly improved. In contrast, when the spectral recovery network is equipped with the physically-constrained RGB transformation network, Network 2 achieves up to \SI{0.75}{\pp}  improvement in Pixel Acc and \SI{1.60}{\pp} improvement in Mean Acc\footnote{p.p. means percentage point}. However, the training process does not guarantee an increase in network performance, as evidenced by the results for filter number 8.

\subsubsection{Domain Alignment.} The domain alignment strategy is designed to learn generalisable features for material datasets without training samples. The reported evaluations show that network 3 improves network performance for all filter choices, indicating that the features extracted by the spectral recovery network are well-tuned for material datasets.

\subsubsection{Multi-Modal Fusion.} The spectradb dataset provides measurements and observations that can help reduce network uncertainty in identifying materials. As shown in Table \ref{tune_filter}, the \netname{} achieves the best performance with an improvement of \SI{0.78}{\pp}/\SI{2.11}{\pp} when the filter number is set to 12 compared with Network 3. The significant increase indicates that human observations, such as roughness, can reliably differentiate between different materials. 

\section{Conclusion}
In this paper, we introduce the \netname{}, a model which employs a physically-constrained spectral recovery network to segment materials using reconstructed hyperspectral images. To leverage the existing spectral recovery dataset, we use domain-aware discriminators to align the material dataset and enhance the quality of the reconstructed hyperspectral image. We also incorporate material observations, such as roughness, to improve the reliability of material predictions. Our experiments demonstrate that the proposed \netname{} outperforms existing models and can handle images captured under varying lighting conditions. The limitation is that the simplified camera model $R(h)$ assumes uniform illumination, while in practice, the indoor illumination can be the combination of multiple light sources, as shown in Section C.3. in the supplementary material. 
\bibliography{aaai24}

\end{document}